\documentclass[10pt,twocolumn,letterpaper]{article}

\usepackage{cvpr}              
\definecolor{cvprblue}{rgb}{0.21,0.49,0.74}
\usepackage[pagebackref,breaklinks,colorlinks,allcolors=cvprblue]{hyperref}
\usepackage{tcolorbox}

\newcommand{\datasetname}{Video2Reaction}

\def\paperID{*****} 
\def\confName{CVPR}
\def\confYear{2026}

\title{\datasetname{}: Training Foundation Video Models \\ to Predict Audience Reaction}

\author{Sidong Zhang\thanks{Equal contribution} \\
UMass Amherst\\
130 Governors Drive, Amherst\\
{\tt\small sidongzhang@umass.edu}
\and
Trang Nguyen\footnotemark[1] \\
UMass Amherst\\
130 Governors Drive, Amherst\\
{\tt\small tramnguyen@cs.umass.edu}
\and
Shiv Shankar \\
UMass Amherst\\
130 Governors Drive, Amherst\\
{\tt\small sshankar@umass.edu}
\and
Gauri Jagatap \\
Dolby Laboratories \\
1275 Market Street, San Francisco\\
{\tt\small Gauri.Jagatap@dolby.com}
\and
Deepak Chandran \\
Dolby Laboratories \\
1275 Market Street, San Francisco\\
{\tt\small deepak.chandran@dolby.com}
\and
Andrea Fanelli  \\
Dolby Laboratories \\
1275 Market Street, San Francisco\\
{\tt\small andrea.fanelli@dolby.com}
\and
Madalina Fiterau \\
UMass Amherst\\
130 Governors Drive, Amherst\\
{\tt\small mfiterau@cs.umass.edu}
}

\begin{document}
\maketitle


\begin{abstract}
    We introduce \textbf{Video2Reaction}, a multimodal dataset that maps short movie segments to the induced emotional reactions of viewers in the wild, as expressed through social media comments. \textbf{Video2Reaction} captures the natural diversity of emotional responses by aggregating reactions from online comments at scale, modeling labels as distributions over categorical emotions to better reflect the subjective and ambiguous nature of emotional perception. We benchmark two vision-language models (VLMs) finetuned with LoRA, showing that VLMs learn effectively from \textbf{Video2Reaction} and outperform specialized baselines on dominant reaction prediction. We further demonstrate that VLMs pre-finetuned on \textbf{Video2Reaction} transfer effectively to VCE, another induced emotion dataset with a different taxonomy and video domain. Notably, LLaVA-NeXT-Video-7B pre-finetuned on \textbf{Video2Reaction} and adapted on only 1\% of VCE training data achieves a top-3 accuracy of $0.682$, on par with the best reported VCE performance trained on the full dataset. The dataset is available at \url{https://huggingface.co/datasets/infofusionlab/Video2Reaction}.
\end{abstract}

\section{Introduction}


Training foundation models to anticipate how audiences react to video content is a crucial yet underexplored challenge in affective computing. If such capability were achieved, it would enable pre-screening of media and iterative content refinement well before public release. However, two major barriers hinder progress in this direction.

First, there is a lack of large-scale data and standardized evaluation for \textit{induced emotion}. Existing video affective datasets~\cite{busso2008iemocap,poria2018meld,zadeh2018multimodal} predominantly focus on \textit{perceived emotions}—for example, the emotions expressed by characters or scenes—while relatively few datasets target \textit{induced emotions}~\cite{tian2017recognizing}, i.e., the emotional responses elicited in viewers. Induced emotions are inherently more complex and variable, as they depend on individual, cultural, and temporal factors.

Second, induced emotions are dynamic and may shift over time, requiring datasets to support continuous updates to remain relevant. Existing approaches~\cite{mazeika2022would, koelstra2011deap, zlatintsi2017cognimuse} typically rely on controlled data collection through recruited participants, capturing audience reactions at a fixed point in time. This process is costly, slow, and difficult to scale.

To address these challenges, we introduce \textbf{\datasetname}, a large-scale multimodal dataset for training foundation models to predict audience reactions to short movie segments. Our dataset leverages social media as a proxy for real-world audience response, aggregating nearly one million YouTube comments across approximately 10,000 videos. We propose a scalable, multi-agent LLM annotation pipeline that extracts structured emotion distributions from raw comments through automated reasoning. Each data instance consists of a video clip paired with a probability distribution over audience reactions, enabling fine-grained modeling of diverse viewer responses. Because annotations are generated automatically, the dataset supports rapid and incremental expansion, allowing it to evolve alongside changing audience preferences.

Empirically, we demonstrate that finetuning two VLMs — Qwen2.5-VL-7B-Instruct and LLaVA-NeXT-Video-7B — on \datasetname\ leads to substantial improvements in audience reaction prediction. Moreover, the resulting models exhibit promising generalization to out-of-distribution datasets under limited supervision. 

\section{\datasetname\ Dataset}

\begin{table}[ht]
\centering
\caption{Finer-grained reaction categories (21) used in \textbf{\datasetname{}} grouped by sentiment.}
\resizebox{\columnwidth}{!}{%
\begin{tabular}{lp{5cm}}
\toprule
\textbf{Sentiment Category} & \textbf{Finer-grained Reaction Categories} \\
\midrule
Positive & amusement, excitement, joy, caring, admiration, relief, approval \\
\midrule
Negative & fear, nervousness, embarrassment, disappointment, sadness, grief, disgust, anger, annoyance, disapproval \\
\midrule
Ambiguous & realization, surprise, curiosity, confusion \\
\bottomrule
\end{tabular}
}
\label{tab:sentiment_mapping}
\end{table}

\begin{figure*}
    \centering
    \includegraphics[width=\linewidth]{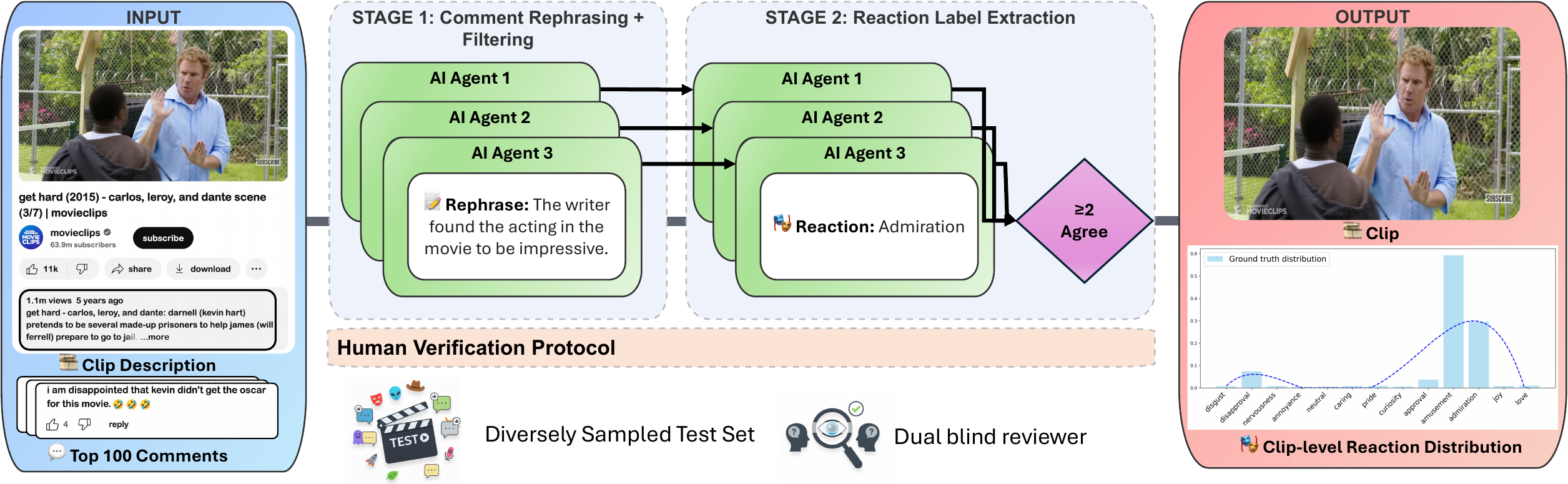}
    \caption{\textbf{Overview of \datasetname{} Two-Stage LLM-based Data Annotation Pipeline.} \textit{Stage 1} rephrases comments to explicitly state their reactions towards the clip. It also filters out comments that lack a discernible reaction to the clip. \textit{Stage 2} extracts reaction labels, with majority voting across three LLM agents to ensure consistency and discard ambiguous cases.
} 
    \label{fig:data_annotation_pipeline}
\end{figure*}

\subsection{Data Collection} \label{sec:data_collection}
\vspace{-0.5em}
We curate movie clips from the CondensedMovies dataset~\citep{bain2020condensed} (licensed with CC BY 4.0), which contains licensed content from the \textit{Movieclips}\footnote{\url{https://www.youtube.com/@MOVIECLIPS}} YouTube channel. The use of licensed content improves the longevity of the dataset, as these clips are less likely to be removed from the platform. To ensure meaningful audience engagement, we retain only videos with a minimum of 10,000 views and at least 10 comments. The selected clips, originally uploaded between 2011 and 2019, are downloaded for further processing. Viewer comments on these videos extend through 2025, resulting in each clip having a minimum of six years of audience commentary. 







\subsection{Reaction Distribution Annotation} 
\label{sec:annotation_pipeline}
\vspace{-0.5em}
\textbf{Reaction Taxonomy.}
To represent the complexity of audience reaction, we adopt the 28-category emotion taxonomy from GoEmotions \citep{demszky2020goemotions}, originally designed for Reddit comments but directly applicable to YouTube, as both platforms share a similar social media language register. However, we drop 7 of the original categories in GoEmotions due to their significant under-representation in our data. These 7 reactions contribute to less than 0.01\% of the distribution mass on average. Our final taxonomy consists of 21 fine-grained emotions (Table \ref{tab:sentiment_mapping}).  

\textbf{Two-stage Multi-agent Reaction Annotation Pipeline.} Given the volume of raw comments and the non-stationary nature of induced emotions over time, we designed a scalable and reliable multi-agent annotation pipeline that supports frequent dataset updates, ensuring long-term relevance and impact. Figure~\ref{fig:data_annotation_pipeline} outlines our LLM-based pipeline for annotating audience reactions based on user comments. Each comment is processed in two stages. The first stage of rephrasing comments is critical, as many audience comments reflect implicit reactions or off-topic remarks that need contextual interpretation to reveal their emotional intent. For example, a comment like \textit{“I’m so disappointed this actor didn’t win an Oscar”} will be mistakenly labeled as \textit{disappointment} without being rephrased as \textit{"the acting in the movie is so impressive"} in the first stage. The second stage then extracts relevant reaction labels from the rephrased comments from the first stage. Prompt details for both stages and qualitative comparison of single-stage versus two-stage annotations are provided in Appendix~\ref{apx:data_annotation}. 

Motivated by prior findings in the use of multi-agent framework in text classification \cite{trad2024ensemble}, compound systems \cite{chen2024more}, and chain-of-thought reasoning \cite{wangself, choi2024multi}, we employ an ensemble of of three medium-sized multilingual instruction-tuned LLMs \footnote{LLaMA-3.1-8B-Instruct \cite{llama3herdmodels}
, Qwen2.5-14B-Instruct \cite{qwen2.5}
, and DeepSeek-R1-Distill-Qwen-7B \cite{deepseekai2025}} and adopt a straightforward majority voting approach for our emotion annotation task. This ensemble design enables inference-time speedup through agent parallelization, in contrast to annotator–critic architectures that require sequential interaction. Our choice to use three medium-sized LLMs with comparable performance aligns with prior observations that ensemble methods are most effective when constituent models exhibit similar strength \cite{trad2024ensemble}.

\textbf{Quality of Reaction Annotations.}
To assess annotation quality, we conduct two complementary evaluations.

\textit{Human–LLM alignment.} We collect independent annotations from 29 participants on 233 video–comment pairs (median 3 annotators per comment) and compute Spearman correlation per emotion following GoEmotions \cite{demszky2020goemotions}. The mean inter-rater correlation is 0.428 (std = 0.233), reflecting the subjective nature of induced emotion labeling. Our LLM pipeline achieves a comparable mean correlation of 0.402 (std = 0.243), operating within the same reliability range as human annotators. Full breakdown by emotion is in Appendix~\ref{sec:error_analysis}.

\textit{Dual-blind verification.} We randomly sample 100 clips across all genres, selecting 10 comments each (1,000 total), independently reviewed by two annotators with a third consulted on disagreements. Overall, 86\% of LLM-assigned labels were judged correct, 7.8\% incorrect, and 6.2\% indeterminate (Table~\ref{tab:reaction_annotation_quality}). A control of randomly-labeled comments yields only 0.2\% correct, ruling out confirmation bias. Performance remains above 70\% across all genres, with Comedy and Drama posing greater challenges due to subtle cues such as pop culture references (Table~\ref{tab:error_by_genre}).



\subsection{Dataset Statistics} \label{sec:data_stats}

The dataset spans 390 hours from 1,545 movies, 10,348 clips (7,243/1,035/2,070 train/val/test), with an average of 44 scenes and 24 comments per clip. \datasetname{} captures substantial variation in audience reactions within the same movie, with a median Chebyshev distance of $0.48$ (ranging from $0$ to $1.0$) between clip-level distributions of the same movie, highlighting the need for clip-level rather than movie-level labeling. The imbalanced factor $\gamma$ is 28.36 across all 21 taxonomies, while the dominant emotion has a median probability of only 0.4, underscoring the importance of modeling full reaction distributions instead of only the dominant label. Details are included in Appendix \ref{sec:data_stats}.

\section{Methods}

\subsection{Problem Setup}
We frame audience reaction prediction as a label distribution learning (LDL) task. LDL predicts a probability distribution over labels, capturing the ambiguity and diversity of audience reactions. Let $x$ denote an input video clip. Its audience reaction is represented by a label distribution $\mathbf{d}_x = \{d_{xm}\}_{m=1}^M$, where $M$ is the number of reaction classes, $d_{xm} \in [0, 1]$ indicates the proportion of viewers associating label $m$ with $x$, and $\sum_{m=1}^{M} d_{xm} = 1$. This can be interpreted as the conditional probability $p(m|x)$. Our objective is to learn a model $f_\theta(x)$ that predicts $\hat{\mathbf{d}}_x \approx \mathbf{d}_x$ by minimizing the KL divergence between $\hat{\mathbf{d}}_x$ and $\mathbf{d}_x$. 

\subsection{Finetuning VLMs to Predict Audience Reaction}
We study \datasetname{} for foundation model training through two experiments. First, we evaluate in-domain VLM finetuning, showing that VLMs can effectively learn audience reaction prediction from \datasetname{} and achieve strong performance on test samples. Second, we leverage the finetuned VLMs for out-of-domain transfer to VCE~\cite{mazeika2022would}, a large-scale induced emotion dataset collected from social media short videos. VCE is a well-motivated transfer target for three reasons: (1) it shares the same induced emotion paradigm as \datasetname{}, making the task semantically close; (2) its videos are sourced from social media while emotions are annotated under controlled conditions, introducing distributional shift in both the visual modality and the label space; and (3) its large scale requires substantial human annotation effort, making it expensive to extend and a natural testbed for evaluating whether \datasetname{} can reduce target-domain annotation cost. We show that the knowledge acquired from \datasetname{} generalizes effectively to VCE, reducing the need for costly target-domain supervision.

\subsection{In-Domain Finetuning of VLMs}

We finetune two VLMs — LLaVA-NeXT-Video-7B~\cite{zhang2024llavanextvideo} and Qwen2.5-VL-7B-Instruct~\cite{qwen2.5-VL} — on \datasetname{} using LoRA of rank 8 applied to attention layers, yielding approximately 6\% trainable parameters. We instruct each VLM to reason about the most likely induced emotion given the input video and clip description, and use the next-token probability over candidate emotion labels as the predicted distribution. This requires that all candidate emotions have distinct first tokens to avoid ambiguity. We adopt two strategies to ensure this: (1) \textit{Label-option}: assigning each emotion a unique letter label and prompting the VLM to respond with only the letter; (2) \textit{Word-option}: rewording any colliding emotion names so their first tokens are non-overlapping, allowing the VLM to reason on the emotion word directly while still enabling unambiguous probability extraction via unique first tokens. Three emotion words are reworded in the Word-option strategy:  disgust is changed to grossed, disapproval to opposed, and disappointment to deflated.

\subsection{Out-of-Domain Transfer Learning of VLMs}
We hypothesize that VLMs finetuned on \datasetname{} acquire transferable knowledge that benefits adaptation to video emotion datasets with shifted video content and label taxonomy. To test this, we compare two settings on VCE~\cite{mazeika2022would}: (1) a VLM pre-finetuned on \datasetname{} and then adapted with a small subset of VCE training data, and (2) a VLM finetuned directly on the same subset of VCE from scratch. Since VCE relies on costly human annotation, a key question is whether \datasetname{} pre-finetuning can reduce the required amount of target-domain supervision while maintaining competitive performance. We evaluate both settings across varying VCE training subset sizes (zero-shot, 500, 1,000, and 5,000 samples), where each subset is drawn via stratified sampling to ensure balanced dominant class representation. Following the in-domain finetuning procedure, we instruct VLMs to respond with the most likely emotion given the video and apply KL divergence between the next single token probabilities and the human-annotated distribution as the training objective. The Word-option strategy requires the following rewording of VCE emotion labels for unique first tokens: adoration to cherish, aesthetic appreciation to beauty, awe (or wonder) to wonder, and anxiety to worry.


\subsection{Baseline Methods}
For in-domain evaluation, we include classic LDL models, LDSVR~\cite{geng2015pre} and SA-BFGS~\cite{gengLabelDistributionLearning2016}, trained from scratch, and zero-shot Gemini 2.5 Flash as a proprietary VLM reference. For out-of-domain transfer, we refer to the state-of-the-art result reported in VCE: VideoMAE trained on the full 50,000 samples with top-3 accuracy $0.689$~\cite{mazeika2022would}.

\section{Results}

\subsection{In-Domain Results}
We evaluate dominant reaction prediction using three metrics: Mean Reciprocal Rank (MRR), which measures the ranking quality of the dominant reaction; Top-1 Probability Error (TPE), which captures probability misestimation of the dominant reaction; and class-weighted Top-$k$ F1 ($F1_k$).

\begin{table}[htb]
\centering
\caption{Dominant reaction evaluation benchmark results. Best performance is shown in bold.}
\resizebox{\columnwidth}{!}{%
\begin{tabular}{lcccc}
\toprule
\textbf{Model Name} & TPE $\downarrow$ & MRR $\uparrow$ & F1$_1$ $\uparrow$ & F1$_3$ $\uparrow$ \\
\midrule
LDSVR 
& 0.1599 & 0.7054 & 0.5034 & 0.5696 \\
SA-BFGS 
& 0.1882 & 0.7163 & 0.5283 & 0.6265 \\
Gemini 2.5 Flash & 0.3026 & 0.4378 & 0.2735 & 0.3794 \\
\midrule
\multicolumn{5}{l}{LLaVA-NeXT-Video-7B} \\
- \textit{Temperature-scaled} 
& 0.4103 & 0.1992 & 0.0143 & 0.1374 \\
- \textit{Finetuned with Label-option} 
& 0.1742 & 0.7632 & 0.5938 & 0.6281 \\
- \textit{Finetuned with Word-option} 
& 0.1643 & 0.7723 & 0.6035 & \textbf{0.6591} \\
\multicolumn{5}{l}{Qwen2.5-VL-7B-Instruct} \\
- \textit{Temperature-scaled} 
& 0.3923 & 0.3088 & 0.1958 & 0.3037 \\
- \textit{Finetuned with Label-option} 
& 0.1692 & 0.7558 & 0.5814 & 0.6386 \\
- \textit{Finetuned with Word-option} 
& \textbf{0.1588} & \textbf{0.7742} & \textbf{0.6158} & 0.6583 \\
\bottomrule
\end{tabular}%
}
\label{tab:dominant_eval}
\end{table}

Table~\ref{tab:dominant_eval} summarizes the dominant reaction evaluation results. Both zero-shot temperature-scaled VLMs and Gemini perform poorly across all metrics, confirming that \textbf{audience reaction prediction requires task-specific adaptation}. After LoRA finetuning, both VLMs substantially outperform the LDL baselines LDSVR and SA-BFGS across most metrics, demonstrating that \datasetname{} provides sufficient and learnable supervision signal for VLMs. Word-option consistently outperforms Label-option for both models, suggesting that reasoning directly over emotion words rather than abstract letter labels better leverages the semantic knowledge encoded in pretrained VLMs.

\subsection{Out-of-Domain Transferable Results}

We adopt the Word-option finetuning strategy based on its superior in-domain performance. We report top-3 accuracy following the evaluation protocol of the original VCE paper~\cite{mazeika2022would}, where the best reported performance is $0.689$, achieved by training VideoMAE on the full 50,000 training samples. Table~\ref{tab:vce_transfer} reports top-3 accuracy for both VLMs with and without \datasetname{} pre-finetuning across varying VCE training subset sizes. In the zero-shot setting, \datasetname{} pre-finetuning alone substantially improves performance over non-pre-finetuned counterparts, demonstrating that reaction knowledge learned from \datasetname{} transfers directly without any VCE supervision. Across all subset sizes, \datasetname{} pre-finetuned VLMs consistently outperform their counterparts trained from scratch on VCE. Notably, LLaVA-NeXT-Video-7B + V2R finetuned on only 500 VCE samples ($0.682$) already matches LLaVA trained on 5,000 samples ($0.701$) and reported SOTA ($0.689$), suggesting that \datasetname{} pre-finetuning reduces the required target-domain annotation by up to an order of magnitude.

\begin{table}[h]
\centering
\caption{Top-3 accuracy on VCE dataset. VLMs finetuned on V2R are on par with full-set trained models with only 1\% of training samples.}
\resizebox{\columnwidth}{!}{%
\begin{tabular}{lcccc}
\toprule
& \multicolumn{4}{c}{\textbf{Size of VCE Training Subset}} \\
\cmidrule(lr){2-5}
\textbf{Model} & \textbf{Zero Shot} & \textbf{500} & \textbf{1000} & \textbf{5000} \\
\textit{VCE Training Subset Ratio} & 0\% & 1\% & 2\% & 10\% \\
\midrule
LLaVA-NeXT-Video-7B          & 0.232 & 0.653 & 0.663 & 0.701 \\
LLaVA-NeXT-Video-7B + V2R    & 0.347 & 0.682 & 0.678 & 0.701 \\
\midrule
Qwen2.5-VL-7B-Instruct          & 0.083 & 0.636 & 0.665 & 0.713 \\
Qwen2.5-VL-7B-Instruct + V2R    & 0.352 & 0.665 & 0.686 & \textbf{0.722} \\
\midrule
\multicolumn{5}{c}{\textbf{Full-Set Training / Baselines}} \\
\midrule
VideoMAE (Full 50K)    & \multicolumn{4}{c}{0.689} \\
Majority Emotion       & \multicolumn{4}{c}{0.357} \\
Random Chance          & \multicolumn{4}{c}{0.111} \\
\bottomrule
\end{tabular}
}
\label{tab:vce_transfer}
\end{table}

\section{Next Steps}

\paragraph{Transfer annotation knowledge for automatic labeling.} We are experimenting with using \datasetname{} pre-finetuned VLMs to automatically label VCE data, then training VideoMAE on the full VLM-labeled 50,000 samples to measure the quality gap against human annotation. Sufficiently high annotation quality would enable scalable, low-cost curation of induced emotion datasets.

\vspace{-4mm}
\paragraph{Understand the effect of model scale on knowledge transfer.} Current experiments are limited to 7B-scale VLMs. We plan to evaluate compact and larger variants of LLaVA and Qwen to understand whether \datasetname{} pre-finetuning benefits scale proportionally or exhibits diminishing returns.

{
    \small
    \bibliographystyle{ieeenat_fullname}
    \bibliography{main}
}


\newpage
\appendix

\section{Data Annotation Pipeline}
\subsection{Implementation Details}
\label{apx:data_annotation}
Given the volume of raw comments to process, we employ an ensemble of three medium-sized multilingual instruction-tuned LLMs—LLaMA-3.1-8B-Instruct\footnote{\url{https://huggingface.co/meta-llama/Llama-3.1-8B-Instruct}}, Qwen2.5-14B-Instruct\footnote{\url{https://huggingface.co/Qwen/Qwen2.5-14B-Instruct}}, and DeepSeek-R1-Distill-Qwen-7B\footnote{\url{https://huggingface.co/deepseek-ai/DeepSeek-R1-Distill-Qwen-7B}}—chosen for their strong performance and favorable efficiency. All LLM agents share the same prompt for both stages, which are listed below.

The LLM annotation pipeline requires two following inputs:
\begin{itemize}
    \item Clip Description, to set context to understand the sentiment of the comments. Our pipeline uses the short clip description provided by @MOVIECLIPS Youtube channel but we can also use a description generated by a video understanding model if no existing description is available.

    \item Comment, user-written comments on youtube.
\end{itemize}

\begin{tcolorbox}[colback=gray!10!white, colframe=gray!30!black, title=Stage 1: Rephrase and Filter Comment Prompt]
You are to roleplay as a senior director speaking to a junior director in training. You are reviewing comments from audience members on a variety of scenes from a variety of movies/shows. You are explaining to the junior director what the audience member is likely feeling due to the clip along with your reasoning. The goal is to teach the junior director how film can predictably be used to invoke certain emotions; as a result, you should ignore comments from audience members where the analysis shows there is likely nothing the junior director can generalize.  
Be concise in explanations, limit them to 2 sentences at most.

\textbf{Example 1:}  
\texttt{<description>} Paul makes a pair of thieves pay for bringing a knife to a gun fight.  
\texttt{<comment>} TTC Subways, this is Toronto these days.  
\texttt{<explanation>} The audience member here is comparing how Toronto subways seems similar to the subways in the scene. You cannot generalize the feelings of this audience member broadly so we will ignore this comment.  
\texttt{<rephrased>} None  

\textbf{Example 2:}  
\texttt{<description>} Crocodile Dundee interrogates a gangster off the side of a building.  
\texttt{<comment>} It's Milton from Office Space.  
\texttt{<explanation>} The audience member just realizes the same actor from another TV show so there is no reaction towards any aspect of the clip here.  
\texttt{<rephrased>} None  

\textbf{Example 3:}  
\texttt{<description>} Marius mourns his fallen comrades.  
\texttt{<comment>} My favorite song of this masterpiece 
\texttt{<explanation>} The audience member likes the music background in the movie clip so this is a reaction we want to know so that we can pay more attention to music and sound in the future.  
\texttt{<rephrased>} The writer loves the music background.  

\textbf{Example 4:}  
\texttt{<description>} The Thénardiers swindle guests at their inn.  
\texttt{<comment>} The 1985 version is the best one yet  
\texttt{<explanation>} The audience member prefers another version of the movie but the comment does not explain why so we will ignore this comment.  
\texttt{<rephrased>} None  

\textbf{Example 5:}  
\texttt{<description>} Walking alone at night, Paul comes face-to-face with an armed criminal.  
\texttt{<comment>} Keep shooting til he's dead, leave no "victim" to identify you later..and sue you..  
\texttt{<explanation>} The audience member reiterates a character's action in the movie, implying that it is a good idea so this is an implicit reaction towards the character's decision or part of the plot in the scene.  
\texttt{<rephrased>} The writer agrees with the character's action in the scene.  

Now, analyze the following comment:  
\texttt{<description>} \{summary\}  
\texttt{<comment>} \{comment\}  
Director:
\end{tcolorbox}

\begin{tcolorbox}[colback=gray!10!white, colframe=gray!30!black, title=Stage 2: Extract Reaction Labels Prompt]
You are an assistant analyzing YouTube comments to extract audience reactions to a movie clip. Given the following inputs:  
- \textbf{clip\_description}: A short description of the movie clip.  
- \textbf{rephrased\_comment}: The original comment rewritten from a third-person perspective.  

Return a JSON object with the following fields:

\begin{itemize}
\item \texttt{high\_level\_reaction}: One or more words from \{"joy", "sadness", "anger", "surprise", "disgust", "fear", "neutral"\}.
\item \texttt{finer\_grained\_reaction}: One or more words from \{"admiration", "amusement", "anger", "annoyance", "approval", "caring", "confusion", "curiosity", "desire", "disappointment", "disapproval", "disgust", "embarrassment", "excitement", "fear", "gratitude", "grief", "joy", "love", "nervousness", "optimism", "pride", "realization", "relief", "remorse", "sadness", "surprise", "neutral"\}.
\item \texttt{reaction\_reason\_type}: One or more from \{"cinematography", "character", "acting", "sound and music", "editing and pacing", "narrative and thematic elements", "personal"\}; if no reason is clear, return \texttt{"none"}.
\end{itemize}

Return only a valid JSON object with these fields and \textbf{no additional text or explanations}.

\textbf{clip\_description}: \{clip\_description\} 

\textbf{rephrased\_comment}: \{rephrased\_comment\}  

\textbf{Output:}
\end{tcolorbox}

\subsection{Human Evaluation \& Additional Error Analysis}
\label{sec:error_analysis}

\textbf{Human-LLM Annotaiton Agreement.} Table \ref{tab:human_eval_emotion_corr} shows a comparison of inter-rater correlation and LLM-Human correlation by 21 emotion classes.

\begin{table}[t]
\centering
\caption{Comparison between human inter-rater agreement and LLM-human correlation across emotion categories.}
\small
\begin{tabular}{lcc}
\hline
\textbf{Emotion} & \textbf{Inter-rater Corr.} & \textbf{LLM-Human Corr.} \\
\hline
admiration        & $0.622 \pm 0.158$  & $0.562 \pm 0.177$ \\
relief            & $0.792 \pm 0.063$  & $0.732 \pm 0.053$ \\
embarrassment     & $0.472 \pm 0.463$  & $0.719 \pm 0.380$ \\
curiosity         & $0.459 \pm 0.301$  & $0.690 \pm 0.281$ \\
confusion         & $0.643 \pm 0.235$  & $0.545 \pm 0.330$ \\
sadness           & $0.847 \pm 0.235$  & $0.524 \pm 0.244$ \\
anger             & $0.511 \pm 0.238$  & $0.484 \pm 0.399$ \\
disapproval       & $0.467 \pm 0.243$  & $0.445 \pm 0.245$ \\
surprise          & $0.279 \pm 0.349$  & $0.439 \pm 0.383$ \\
grief             & $0.740 \pm 0.307$  & $0.437 \pm 0.277$ \\
realization       & $0.292 \pm 0.289$  & $0.381 \pm 0.389$ \\
amusement         & $0.431 \pm 0.288$  & $0.380 \pm 0.290$ \\
joy               & $0.331 \pm 0.253$  & $0.285 \pm 0.232$ \\
excitement        & $0.162 \pm 0.270$  & $0.275 \pm 0.421$ \\
disgust           & $0.434 \pm 0.339$  & $0.265 \pm 0.330$ \\
disappointment    & $0.186 \pm 0.266$  & $0.191 \pm 0.404$ \\
annoyance         & $0.368 \pm 0.289$  & $0.165 \pm 0.236$ \\
approval          & $0.495 \pm 0.126$  & $0.070 \pm 0.176$ \\
fear              & $-0.043 \pm 0.013$ & $0.903 \pm 0.097$ \\
caring            & $-0.042 \pm 0.013$ & $-0.022 \pm 0.000$ \\
nervousness       & $0.541 \pm 0.171$  & $-0.031 \pm 0.006$ \\
\hline
\textbf{Mean} & $\mathbf{0.428}$ & $\mathbf{0.402}$ \\
\hline
\end{tabular}

\label{tab:human_eval_emotion_corr}
\end{table}
\noindent
\textbf{Dual-blind Human Verification.} We provide a summary of human evaluation on our test set in Table \ref{tab:reaction_annotation_quality}. We further analyze different types of errors that our annotation pipeline tend to make and present them in Table \ref{tab:error_by_emotion} and \ref{tab:error_by_genre}.


\begin{table}[ht]
\centering
\caption{Human evaluation of automated reaction annotation on a sample of 1000 comments.}
\small
\begin{tabular}{lp{3cm}l}
\toprule
\textbf{Human Rating} & \textbf{Description} & \textbf{\# comments (\%)} \\
\midrule
Correct     & Most annotators agree the LLM-assigned labels are correct.      & 860 (86.0\%) \\
Incorrect   & Most annotators agree the LLM-assigned labels are incorrect.    & 78 (7.8\%)  \\
Not Sure    & Most annotators are unsure about the correctness of the labels (i.e. due to lack of context on movie references). & 62 (6.2\%)  \\
\bottomrule
\end{tabular}
\label{tab:reaction_annotation_quality}
\end{table}

\begin{table}[ht]
\centering
\caption{Performance across different movie genres.}
\small
\begin{tabular}{lrrr}
\hline
\textbf{Movie Genre} & \textbf{\% Correct} & \textbf{\% Not Sure} & \textbf{\% Incorrect} \\
\hline
Adventure   & 96.23 &  1.89 &  1.89 \\
Fantasy     & 96.23 &  1.89 &  1.89 \\
Film-Noir   & 95.00 &  5.00 &  0.00 \\
Musical     & 93.94 &  0.00 &  6.06 \\
Documentary & 90.48 &  4.76 &  4.76 \\
History     & 90.20 &  3.92 &  5.88 \\
Sci-Fi      & 89.58 &  4.17 &  6.25 \\
Biography   & 88.24 &  5.88 &  5.88 \\
Romance     & 87.27 &  3.64 &  9.09 \\
Crime       & 87.04 &  7.41 &  5.56 \\
Horror      & 86.79 &  3.77 &  9.43 \\
Sport       & 84.09 &  4.55 & 11.36 \\
Thriller    & 83.33 &  1.85 & 14.81 \\
Mystery     & 83.02 &  3.77 & 13.21 \\
Music       & 81.63 & 16.33 &  2.04 \\
Animation   & 80.77 & 11.54 &  7.69 \\
War         & 80.39 &  5.88 & 13.73 \\
Family      & 80.00 &  7.27 & 12.73 \\
Comedy      & 79.25 & 16.98 &  3.77 \\
Drama       & 77.78 & 14.81 &  7.41 \\
Action      & 72.55 & 15.69 & 11.76 \\
\hline
\end{tabular}

\label{tab:error_by_genre}
\end{table}

\begin{table}[ht]
\centering
\caption{Performance across different emotion categories.}
\small
\begin{tabular}{lrrr}
\hline
\textbf{Emotion Category} & \textbf{\% Correct} & \textbf{\% Not Sure} & \textbf{\% Incorrect} \\
\hline
annoyance      & 100.00 &  0.00 &  0.00 \\
relief         & 100.00 &  0.00 &  0.00 \\
confusion      &  94.64 &  5.36 &  0.00 \\
approval       &  93.94 &  0.00 &  6.06 \\
curiosity      &  92.31 &  0.00 &  7.69 \\
amusement      &  91.53 &  5.29 &  3.17 \\
admiration     &  90.41 &  5.17 &  4.43 \\
surprise       &  85.71 &  2.86 & 11.43 \\
disapproval    &  84.71 &  7.85 &  7.44 \\
excitement     &  82.35 & 17.65 &  0.00 \\
disappointment &  73.91 &  8.70 & 17.39 \\
disgust        &  72.73 &  9.09 & 18.18 \\
sadness        &  66.67 & 20.00 & 13.33 \\
fear           &  66.67 & 11.67 & 21.67 \\
realization    &  66.67 &  0.00 & 33.33 \\
joy            &  50.00 & 16.67 & 33.33 \\
caring         &  50.00 & 25.00 & 25.00 \\
grief          &  50.00 & 16.67 & 33.33 \\
anger          &   0.00 & 75.00 & 25.00 \\
embarrassment  &   0.00 & 33.33 & 66.67 \\
nervousness    &   0.00 &  0.00 &100.00 \\
\hline
\end{tabular}
\label{tab:error_by_emotion}
\end{table}

\section{Data Statistics}
\label{sec:data_stats}
\begin{figure*}[ht]
    \centering

    \begin{subfigure}[t]{0.48\linewidth}
        \centering
        \includegraphics[width=\linewidth]{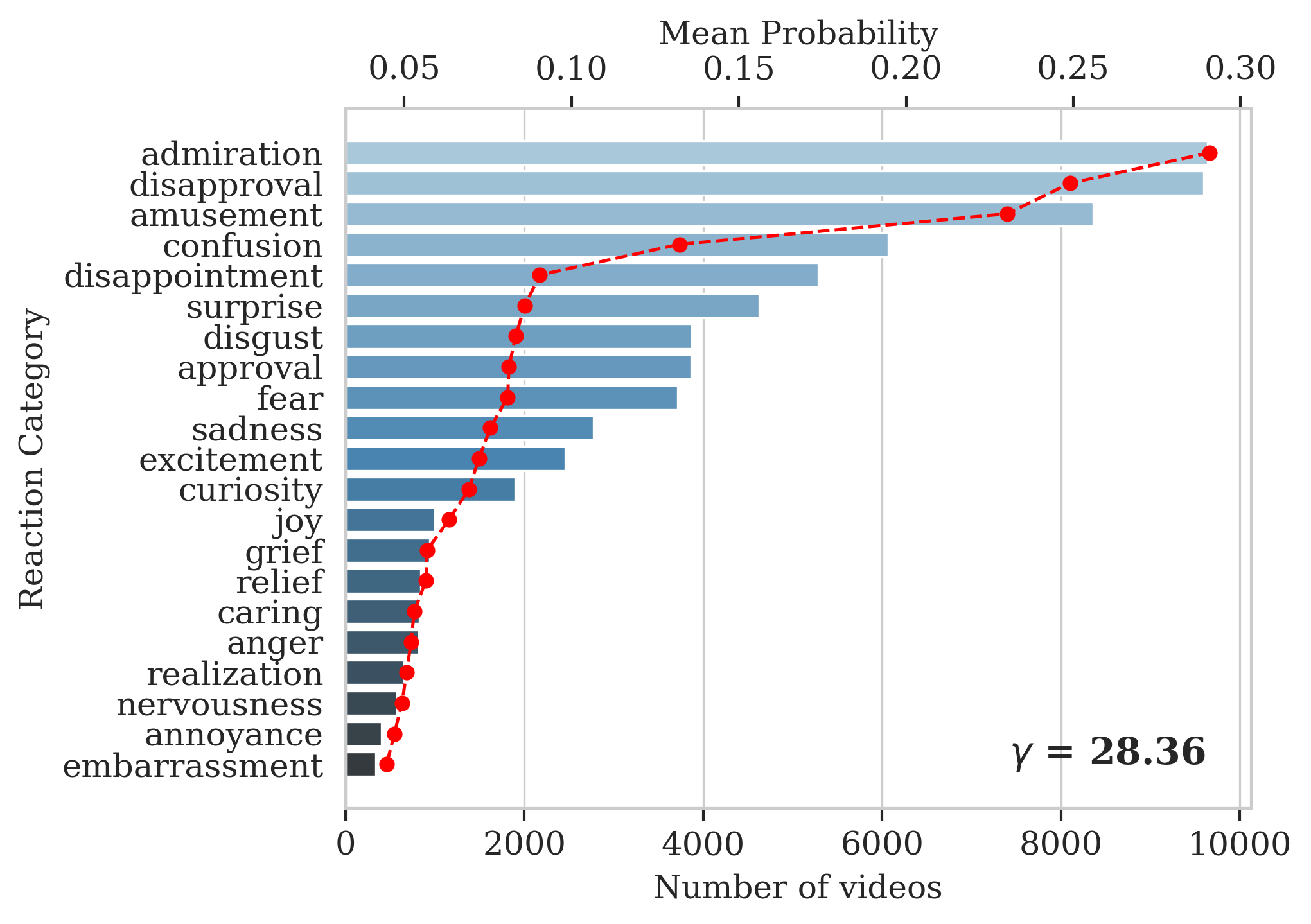}
        \caption{Total number of videos and mean video-level probability of each reaction category ($\gamma$ denotes imbalance factor)}
        \label{fig:reaction_distribution}
    \end{subfigure}
    \hfill
    \begin{subfigure}[t]{0.48\linewidth}
        \centering
        \includegraphics[width=\linewidth]{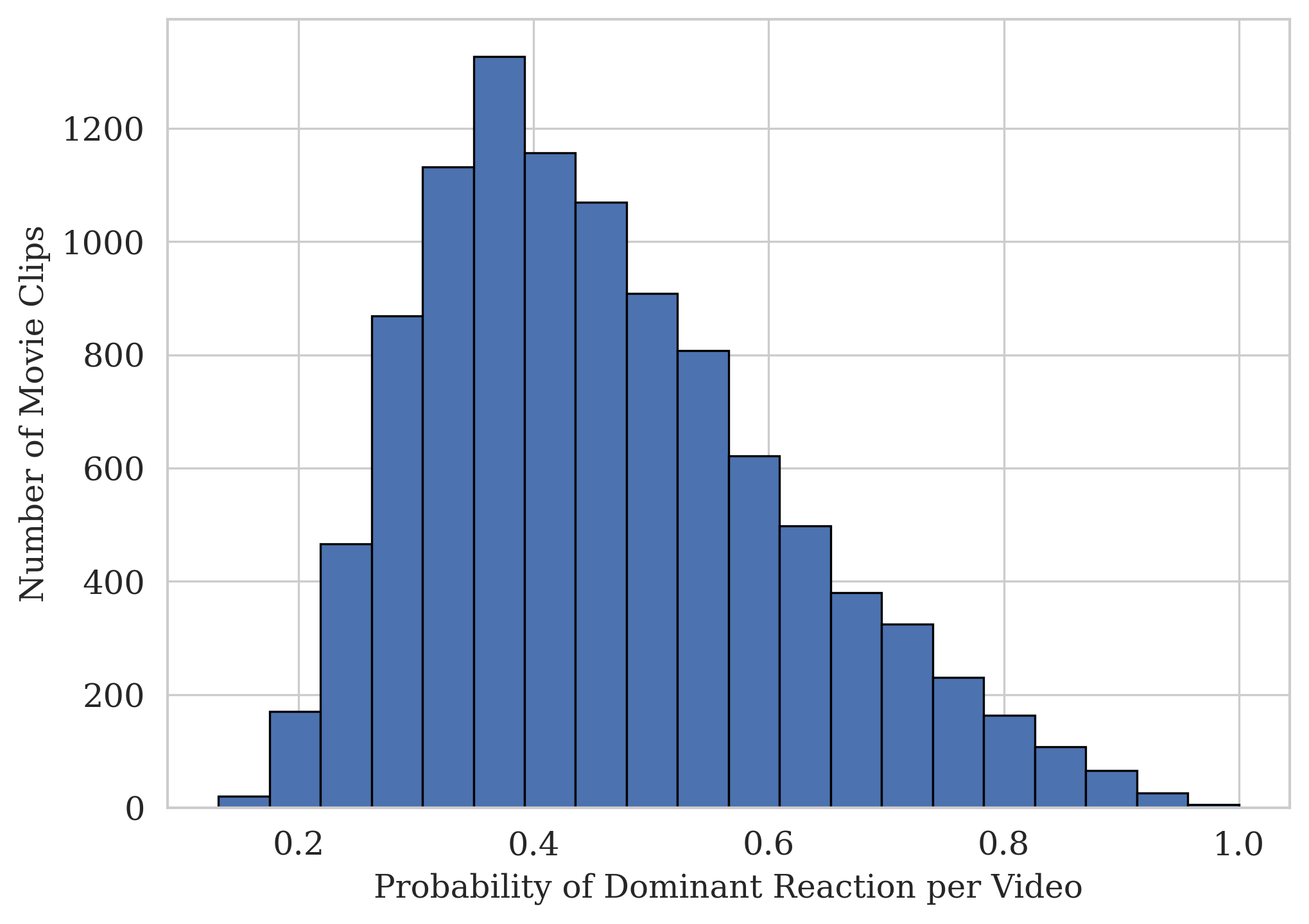}
        \caption{Distribution of dominant reaction probability per video}
        \label{fig:dominant_reaction_prob}
    \end{subfigure}
    \caption{Key Statistics on Reaction Outcome in the Video2Reaction dataset. }
    \label{fig:reaction_stats_combined}
\end{figure*}

Table~\ref{tab:video_comment_stats} summarizes \datasetname{} at the movie, clip, and comment levels. As shown in Figure~\ref{fig:reaction_distribution}, the dataset is highly imbalanced across the 21 reaction categories (with an imbalance factor of $28.36$). Figure~\ref{fig:dominant_reaction_prob} further shows that top-1 reaction probability varies considerably across clips (with a median of approximately $0.4$). 


\begin{table}[ht]
\centering
\caption{Descriptive statistics for video and comment data in \datasetname{}.}
\resizebox{0.5\textwidth}{!}{%
\begin{tabular}{lccccc}
\toprule
\textbf{Category} & \textbf{Total} & \textbf{Min} & \textbf{Mean} & \textbf{Median} & \textbf{Max} \\
\midrule
\multicolumn{6}{l}{\textbf{Movie-level Statistics}} \\
\midrule
Inter-segment Chebyshev distance & 1,545 & 0.05 & 0.48 & 0.48 & 0.91 \\
in reaction distribution ([0.0,1.0]) & & & & & \\
\midrule
\multicolumn{6}{l}{\textbf{Clip-level Statistics}} \\
\midrule
Clip Duration (sec) & 389.81 hrs & 23.15 & 135.61 & 127.60 & 367.36 \\
Key Scenes & 455,226 & 16 & 43.99 & 39.00 & 176 \\
\midrule
\multicolumn{6}{l}{\textbf{Comment-level Statistics}} \\
\midrule
Raw Comments & 771,684 & 19 & 74.57 & 77.00 & 100 \\
Retained Comments & 252,462 & 10 & 24.40 & 22.00 & 100 \\
\bottomrule
\end{tabular}%
}
\label{tab:video_comment_stats}
\end{table}

\end{document}